%% file: main.tex
\documentclass[10pt,twocolumn,letterpaper]{article}

\usepackage{cvpr}              

\usepackage{graphicx}
\usepackage{amsmath}
\usepackage{amssymb}
\usepackage{booktabs}

\usepackage{lipsum}

\usepackage[pagebackref,breaklinks,colorlinks]{hyperref}

\usepackage[capitalize]{cleveref}
\crefname{section}{Sec.}{Secs.}
\Crefname{section}{Section}{Sections}
\Crefname{table}{Table}{Tables}
\crefname{table}{Tab.}{Tabs.}

\def\cvprPaperID{7224} 
\def\confName{CVPR}
\def\confYear{2023}

\begin{document}

\title{GM-NeRF: Learning Generalizable Model-based Neural Radiance Fields \\ from Multi-view Images}


\author{Jianchuan Chen$^1$\footnotemark[1] \quad
Wentao Yi$^1$\footnotemark[1] \quad
Liqian Ma$^2$$^\dag$ \quad
Xu Jia$^1$ \quad
Huchuan Lu$^1$$^\dag$ \\[1.5mm]
$^1$ Dalian University of Technology, China \quad $^2$ ZMO AI Inc.
}

\maketitle
\renewcommand{\thefootnote}{\fnsymbol{footnote}}
\footnotetext[1]{Equal contribution. $^\dag$Corresponding authors. Codes are available at \url{https://github.com/JanaldoChen/GM-NeRF}}

\input{sections/00_abstract}

\input{sections/01_introduction}

\input{sections/02_related_works}

\input{sections/03_method}

\input{sections/04_experiments}

\input{sections/05_limitations}

\input{sections/06_conclusion}

\vspace{1.5em}
\noindent\textbf{Acknowledgements.} 
The paper is supported in part by the National Key R$\&$D Program of China under Grant No. 2018AAA0102001 and the National Natural Science Foundation of China under grant No.62293542, U1903215, and the Fundamental Research Funds for the Central Universities No.DUT22ZD210.


\clearpage

{\small
\bibliographystyle{ieee_fullname}
\bibliography{references}
}

\end{document}

%% file: sections/00_abstract.tex
\begin{abstract}

In this work, we focus on synthesizing high-fidelity novel view images for arbitrary human performers, given a set of sparse multi-view images. 
It is a challenging task due to the large variation among articulated body poses and heavy self-occlusions. 
To alleviate this, we introduce an effective generalizable framework Generalizable Model-based Neural Radiance Fields (GM-NeRF) to synthesize free-viewpoint images. 
Specifically, we propose a geometry-guided attention mechanism to register the appearance code from multi-view 2D images to a geometry proxy which can alleviate the misalignment between inaccurate geometry prior and pixel space. On top of that, we further conduct neural rendering and partial gradient backpropagation for efficient perceptual supervision and improvement of the perceptual quality of synthesis.
To evaluate our method, we conduct experiments on synthesized datasets THuman2.0 and Multi-garment, and real-world datasets Genebody and ZJUMocap. The results demonstrate that our approach outperforms state-of-the-art methods in terms of novel view synthesis and geometric reconstruction.
\end{abstract}

%% file: sections/01_introduction.tex
\vspace{-1em}
\section{Introduction}

3D digital human reconstruction has a wide range of applications in movie production, telepresence, 3D immersive communication, and AR/VR games. Traditional digital human production relies on dense camera arrays~\cite{ARF,LightStage} or depth sensors~\cite{Self-portraits, Fusion4D} followed by complex graphics rendering pipelines for high-quality 3D reconstruction, which limits the availability to the general public. 

\begin{figure}[ht]
    \centering
    \includegraphics[width=1.0\linewidth]{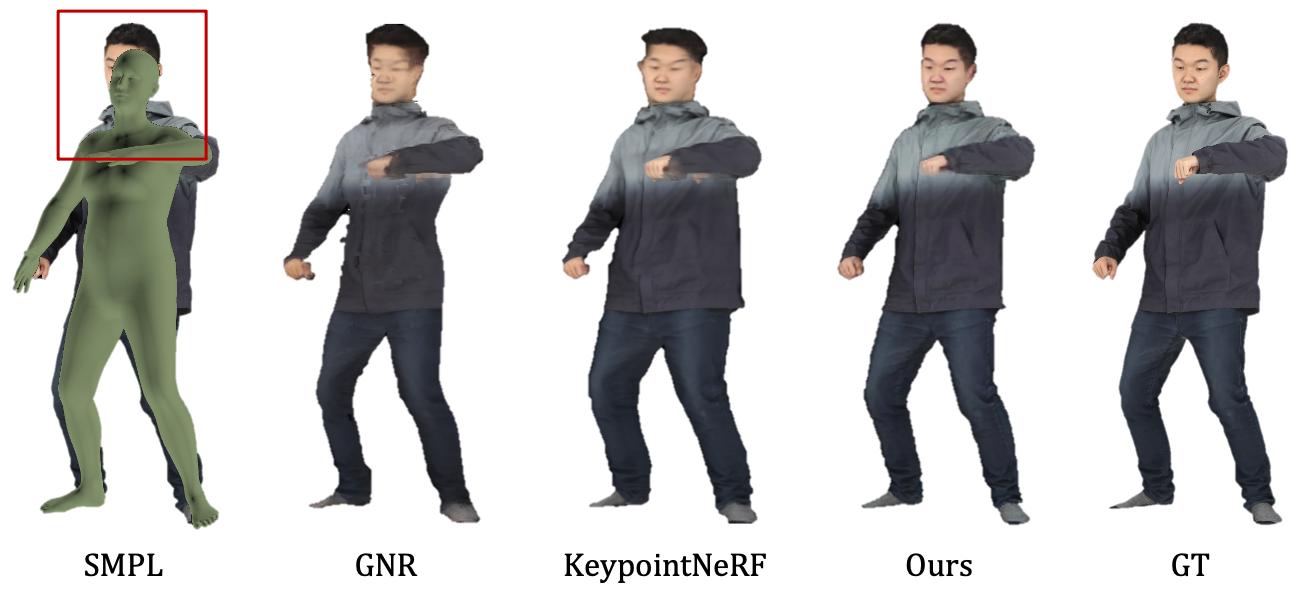}
    \vspace{-2.5em}
    \caption{{\bf The effect of inaccurately estimated SMPL}. Compared with GNR\cite{genebody} and KeypointNeRF\cite{keypointNeRF}, our method  still yields a reasonable result.}
    \vspace{-1.5em}
    \label{fig:teaser}
\end{figure}

Reconstructing 3D humans from 2D images captured by sparse RGB cameras is very attractive due to its low cost and convenience.
This field has been studied for decades~\cite{PCMSA,Scalable3D,DSC}.
However, reconstruction from sparse RGB cameras is still quite challenging because of: 1) heavy self-occlusions of the articulated human body; 2) inconsistent lighting and sensor parameters between different cameras; 3) highly non-rigid and diverse clothes.

In recent years, with the rise of learning-based methods, we can reconstruct high-quality digital humans from sparse cameras. 
Learning-based methods \cite{NT,NHR,TNA,SMPLpix,vid2vid} have made great processes, however, they lack multi-view geometric consistency due to the mere usage of a 2D neural rendering network. 
To address this problem, many recent works\cite{PixelNeRF, ibrnet, MVSNeRF} adopt neural radiance fields as 3D representations, which achieves outstanding performance on novel view synthesis.
However, these methods are not robust to unseen poses without the guidance of human geometric prior.

To better generalize to unseen poses, NeuralBody~\cite{neuralbody} introduces a statistical body model SMPL\cite{SMPL} into neural radiance fields which can reconstruct vivid digital humans from a sparse multi-view video. 
However, NeuralBody is designed for identity-specific scenarios, which means it requires laborious data collection and long training to obtain the model for one person.
Such a limitation restricts its application in general real-world scenarios.

In this work, we focus on synthesizing high-fidelity novel view images for arbitrary human performers from a set of sparse multi-view images.
Towards this goal, some very recent works \cite{NHP, genebody, gpnerf, keypointNeRF} propose to aggregate multi-view pixel-aligned features using SMPL as a geometric prior.
However, these methods usually assume perfect geometry (\eg accurate SMPL~\cite{SMPL} estimation from 2D images) which is not applicable in practical applications. In practice, the geometry error does affect the reconstruction performance significantly. 
As illustrated in the red box of \cref{fig:teaser}, when the estimated SMPL does not align well with RGB image, prior SMPL-dependent methods\cite{genebody, keypointNeRF} yield blurry and distorted results.
The such performance gap is caused by the misalignment between the 3d geometry (\ie SMPL) and the pixel space (\ie pixel-aligned feature and ground-truth image).
Specifically, the misalignment will cause: 1) blur and distortion when fusing the geometry and pixel-aligned features; 2) unsuitable supervision during training with a pixel-wise loss like L1 or L2.
To alleviate the issue of misalignment, we propose to take the geometry code as a proxy and then register the appearance code onto the geometry through a novel geometry-guided attention mechanism.
Furthermore, we leverage perceptual loss to reduce the influence of misalignment and promote sharp image synthesis, which is evaluated at a higher level with a larger perceptual field. 
It is non-trivial to apply perceptual loss in NeRF-based methods as the perceptual loss requires a large patch size as input which is memory-consuming through volume rendering. We introduce 2D neural rendering and partial gradient backpropagation to alleviate the memory requirement and enhance the perceptual quality.

To summarize, our work contributes as follows:

$\bullet$ A novel generalizable model-based framework GM-NeRF is proposed for the free-viewpoint synthesis of arbitrary performers.

$\bullet$ To alleviate the misalignment between 3D geometry and the pixel space, we propose geometry-guided attention to aggregate multi-view appearance and geometry proxy.

$\bullet$ To enable perceptual loss supervision to further alleviate misalignment issues, we adopt several efficient designs including 2D neural rendering and partial gradient backpropagation.


%% file: sections/02_related_works.tex
\section{Related work}
\noindent \textbf{Implict Neural Representation}.
Implicit neural representations (also known as coordinate-based representations) are a popular way to parameterize content of all kinds, such as audio, images, video, or 3D scenes~\cite{FFL, siren, srn, NeRF}.
Recent works \cite{NeRF, DeepSDF, occnet, srn} build neural implicit fields for geometric reconstruction and novel view synthesis achieving outstanding performance.
The implicit neural representation is continuous, resolution-independent, and expressive, and is capable of reconstructing geometric surface details and rendering photo-realistic images. 
While explicit representations like point clouds\cite{points1, NHR}, meshes\cite{NT}, and voxel grids\cite{deepvoxels, occnet, NeuralVolume, voxel1} are usually limited in resolution due to memory and topology restrictions.
One of the most popular implicit representations - Neural Radiance Field (NeRF) \cite{NeRF} -  proposes to combine the neural radiance field with differentiable volume for photo-realistic novel views rendering of static scenes. However, NeRF requires optimizing the 5D neural radiance field for each scene individually, which usually takes hours to converge. Recent works\cite{PixelNeRF, ibrnet, MVSNeRF} try to extend NeRF to generalization with sparse input views.
In this work, we extend the neural radiance field to a general human reconstruction scenario by introducing conditional geometric code and appearance code.

\noindent \textbf{3D Model-based Human Reconstruction}
With the emergence of human parametric models like SMPL\cite{SMPL,SMPLX} and SCAPE\cite{SCAPE}, many model-based 3D human reconstruction works have attracted wide attention from academics. Benefiting from the statistical human prior, some works\cite{tex2shape, Multi-Garment, expose, VIBE} can reconstruct the rough geometry from a single image or video. 
However, limited by the low resolution and fixed topology of statistical models, these methods cannot represent arbitrary body geometry, such as clothing, hair, and other details well. 
To address this problem, some works\cite{PIFu, pifuhd} propose to use pixel-aligned features together with neural implicit fields to represent the 3D human body, but still have poor generalization for unseen poses. To alleviate such generalization issues, \cite{pamir, arch, doublefield} incorporate the human statistical model SMPL\cite{SMPL, SMPLX} into the implicit neural field as a geometric prior, which improves the performance on unseen poses. 
Although these methods have achieved stunning performance on human reconstruction, high-quality 3D scanned meshes are required as supervision, which is expensive to acquire in real scenarios. Therefore, prior works\cite{PIFu, pifuhd, pamir, arch} are usually trained on synthetic datasets and have poor generalizability to real scenarios due to domain gaps. To alleviate this limitation, 
some works\cite{neuralbody, Anim-NeRF, animnerf_zju, humannerf, arah, a-nerf}  combine neural radiance fields\cite{NeRF} with SMPL\cite{SMPL} to represent the human body, which can be rendered to 2D images by differentiable rendering. 
Currently, some works\cite{gpnerf, genebody, NHP, keypointNeRF, doublediffuse, doublefield} can quickly create neural human radiance fields from sparse multi-view images without optimization from scratch.
While these methods usually rely on accurate SMPL estimation which is not always applicable in practical applications. 


\begin{figure*}[ht]
    \centering
    \vspace{-1em}
    \includegraphics[width=1.0\linewidth]{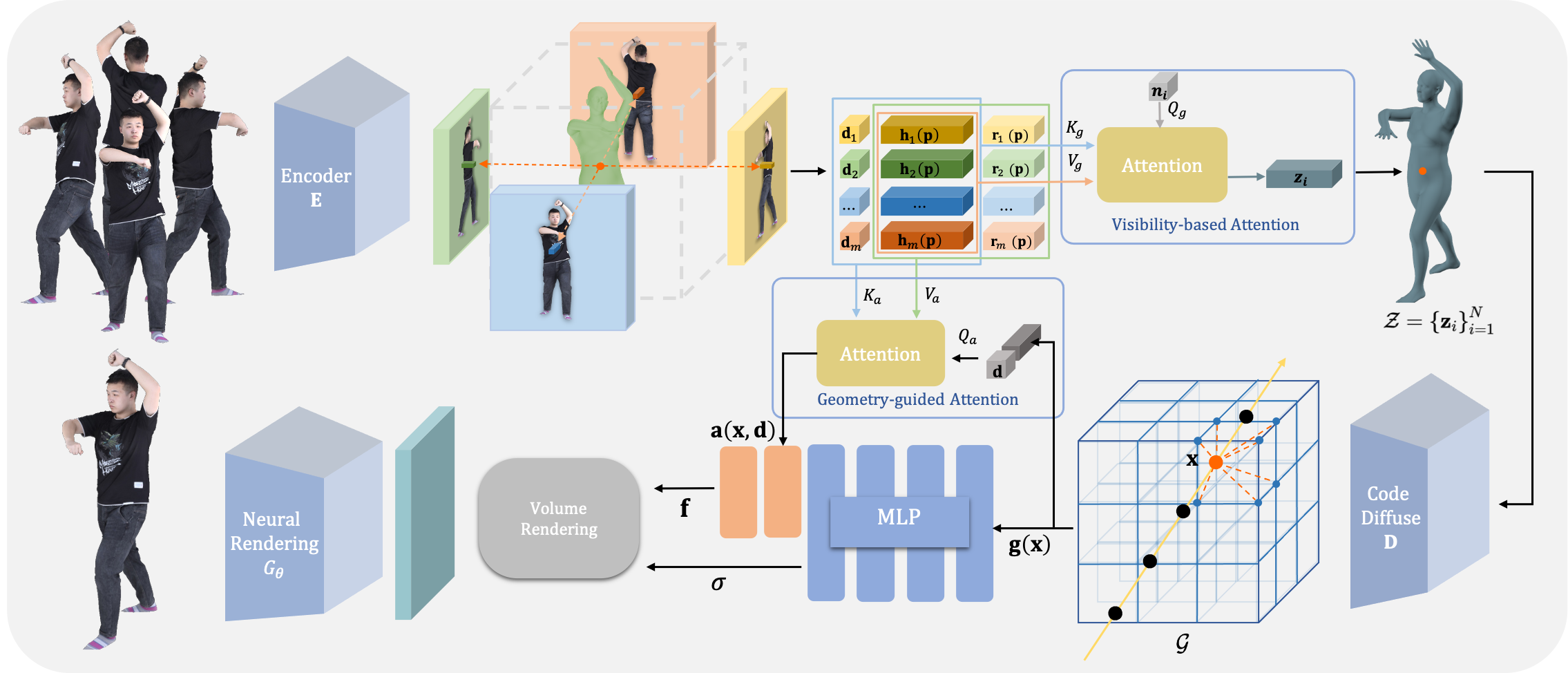}
    \vspace{-1.5em}
    \caption{\textbf{The architecture of our method}. Given $m$ calibrated multi-view images and registered SMPL, we build the generalizable model-based neural human radiance field. First, we utilize the image encoder to extract multi-view image features, which are used to provide geometric and appearance information, respectively. In order to adequately exploit the geometric prior, we propose the visibility-based attention mechanism to construct a structured geometric body embedding, which is further diffused to form a geometric feature volume. For any spatial point $\mathbf{x}$, we trilinearly interpolate the feature volume $\mathcal{G}$ to obtain the geometric code $\mathbf{g}(\mathbf{x})$. In addition, we also propose geometry-guided attention to obtain the appearance code $\mathbf{a}(\mathbf{x}, \mathbf{d})$ directly from the multi-view image features. We then feed the geometric code $\mathbf{g}(\mathbf{x})$ and appearance code $\mathbf{a}(\mathbf{x}, \mathbf{d})$ into the MLP network to build the neural feature field $(\mathbf{f}, \sigma) = F(\mathbf{g}(\mathbf{x}), \mathbf{a}(\mathbf{x}, \mathbf{d}))$. Finally, we employ volume rendering and neural rendering to generate the novel view image.
    }
    \vspace{-1em}
    \label{fig:architecture}
\end{figure*}

%% file: sections/03_method.tex


\section{Method}




We introduce an effective framework GM-NeRF for novel view synthesis and 3D human reconstruction as illustrated in ~\cref{fig:architecture}. GM-NeRF learns generalizable model-based neural radiance fields from calibrated multi-view images by introducing a parametric model SMPL as a geometric prior, which can generalize to unseen identity and unseen pose.

Given $m$ calibrated multi-view images $\left\{I_{k}\right\}_{k=1}^{m}$ of a person, we use Easymocap\cite{EasyMocap} to obtain the SMPL\cite{SMPL} parameters $\mathbf{M}(\theta, \beta)$ of the person. We feed the multi-view images into the encoder network $\mathbf{E}$ to extract multi-view feature maps,
\begin{equation}
\begin{aligned}
H_{k} &= \mathbf{E}(I_{k}), \quad k=1,2,\ldots,m .
\end{aligned}
\end{equation}

For any 3D position $\mathbf{p}$, we can project it onto the feature map $H_{k}$ according to the corresponding camera parameters, which is defined as $\pi _{k}(\cdot)$, then use bilinear interpolation $\Psi(\cdot)$ to obtain the pixel-aligned feature $\mathbf{h}_{k}(\mathbf{p})$ and pixel-aligned color $\mathbf{r}_{k}(\mathbf{p})$ as follows,
\begin{equation}
\begin{aligned}
\mathbf{h}_{k}(\mathbf{p}) &= \Psi (H_{k}, \pi _{k}(\mathbf{p})), \\
\mathbf{r}_{k}(\mathbf{p}) &= \Psi (I_{k}, \pi _{k}(\mathbf{p})).
\end{aligned}
\end{equation}
In order to adequately exploit the geometric prior, we propose the visibility-based attention mechanism to construct a structured geometric body embedding, which is further diffused to form a geometric feature volume (\cref{Structured_Geometric_Body_Embedding}). Afterward, we trilinear interpolate each spatial point $\mathbf{x}$ in the feature volume $\mathcal{G}$ to obtain the geometric code $\mathbf{g}(\mathbf{x})$. 
To avoid the misalignment between the appearance code and geometry code, we utilize the geometry code as a proxy and then register the appearance code $\mathbf{a}(\mathbf{x}, \mathbf{d})$ directly from the multi-view image features with a novel geometry-guided attention mechanism (\cref{Multi-View_Appearance_Blending}).
We then feed the geometric code $\mathbf{g}(\mathbf{x})$ and appearance code $\mathbf{a}(\mathbf{x}, \mathbf{d})$ into the MLP network to build the neural feature field $(\mathbf{f}, \sigma) = F(\mathbf{g}(\mathbf{x}), \mathbf{a}(\mathbf{x}, \mathbf{d}))$ followed by volume rendering and neural rendering for novel view image generation (\cref{Differential_Rendering}). 
To obtain high-quality results, we carefully design an optimization objective including a novel normal regularization. (\cref{Loss_Functions}) as well as an efficient training strategy (\cref{Efficient_Training}).

\subsection{Structured Geometric Body Embedding}
\label{Structured_Geometric_Body_Embedding}
Different from neural radiance fields on general scenes, we introduce a parametric body model to provide the geometric prior for constructing the neural human radiance field, which can enhance generalizability under unseen poses. In our experiments, we choose the SMPL\cite{SMPL} model as the parametric model. The SMPL\cite{SMPL} model $\mathbf{M}(\theta, \beta)$ is a mesh with $N=6,890$ vertices $\left\{\mathbf{v}_{i}\right\}_{i=1}^{N}$, where it is mainly controlled by the pose parameter 
$\theta$, and the shape parameter $\beta$. 
NeuralBody\cite{neuralbody} optimizes a set of structured latent codes from scratch on vertices of the SMPL model for each specific identity. However, not only does it fail to represent a new identity but also has poor generalizability on unseen poses. To address such limitation, we extract the structured latent codes $\mathcal{Z}=\left\{\mathbf{z}_{i}\right\}_{i=1}^{N}$ from the multi-view feature map $H_{k}$ as a geometric embedding to represent arbitrary identities. 
For vertex $\mathbf{v}_{i}$, we design a visibility-based attention mechanism as shown in \cref{fig:architecture} to fuse multi-view features.
\begin{equation}
\begin{aligned}
\mathbf{Q}_{g}(\boldsymbol{v}_{i}) &=F_{Q}^{g}\left(\boldsymbol{n}_{i}\right) \\
\mathbf{K}_{g}(\boldsymbol{v}_{i}) &=F_{K}^{g}(\{\mathbf{h}_{k}(\boldsymbol{v}_{i}) \oplus \mathbf{d}_{k}\}_{k=1}^{m}) \\
\mathbf{V}_{g}(\boldsymbol{v}_{i}) &=F_{V}^{g}(\{\mathbf{h}_{k}(\boldsymbol{v}_{i})\}_{k=1}^{m}) \\
\mathbf{z}_{i} &=F^{g}\left(Att\left(\mathbf{Q}_{g}(\boldsymbol{v}_{i}), \mathbf{K}_{g}(\boldsymbol{v}_{i}), \mathbf{V}_{g}(\boldsymbol{v}_{i})\right)\right)
\end{aligned}
\end{equation}
where $\oplus$ is the concatenation operator, and $\boldsymbol{n}_{i}$ is the normal of the vertex $\boldsymbol{v}_{i}$. $F_{Q}^{g}$, $F_{K}^{g}$, $F_{V}^{g}$ denote the geometric linear layers producing the query, key, and value matrices $\mathbf{Q}_{g}(\boldsymbol{v}_{i})$, $\mathbf{K}_{g}(\boldsymbol{v}_{i})$, $\mathbf{V}_{g}(\boldsymbol{v}_{i})$, respectively. $Att$ is the attention mechanism proposed by \cite{attention}. $F^{g}$ is the geometric feed-forward layer. The intuition of this visibility-based attention mechanism is that the closer the input camera direction $\mathbf{d}_{k}$ is to the normal $\boldsymbol{n}_{i}$, the more the corresponding feature contributes. As shown in \cref{fig:att_vis}, the visualization result demonstrates the plausibility of this design.

Similar to NeuralBody\cite{neuralbody}, we use SparseConvNet\cite{spconv} $\mathbf{D}$ to diffuse the structured latent codes $\{\mathbf{z}_{i}\}_{i=1}^{N}$ into the nearby space to form a 3D feature volume $\mathcal{G}$.

\begin{equation}
\begin{aligned}
    \mathcal{G} = \mathbf{D} (\{\mathbf{z}_{i}\}_{i=1}^{N}) \\
    \mathbf{g}(\mathbf{x}) = \Phi (\mathbf{x}, \mathcal{G} )
\end{aligned}
\end{equation}
where $\Phi(\cdot)$ is the trilinear interpolation operation, which is applied to obtain the geometric code $\mathbf{g}(\mathbf{x})$ for any 3D position $\mathbf{x}$ during volume rendering.

\subsection{Multi-View Appearance Blending}
\label{Multi-View_Appearance_Blending}


Although the structured geometric body embedding provides a robust geometric prior, high-frequency appearance details such as wrinkles and patterns are lost, due to the low resolution and the minimally-clothed topology of the parametric model. 
In practice, inaccurate SMPL estimation will lead to the misalignment between the 3D geometry and pixel space, which will cause blur and distortion when fusing the geometry and pixel-aligned feature.
To solve this problem, we design a geometry-guided attention mechanism as shown in \cref{fig:architecture}, which utilizes the geometry code as a proxy and then registers the appearance code $\mathbf{a}(\mathbf{x}, \mathbf{d})$  directly from the multi-view image features for any 3D position $\mathbf{x}$ and view direction $\mathbf{d}$.
\begin{equation}
\begin{aligned}
\mathbf{Q}_{a}(\mathbf{x}) &=F_{Q}^{a}\left(\mathbf{g}(\mathbf{x}) \oplus \mathbf{d}\right) \\
\mathbf{K}_{a}(\mathbf{x}) &=F_{K}^{a}(\{\mathbf{h}_{k}(\mathbf{x}) \oplus \mathbf{d}_{k}\}_{k=1}^{m}) \\
\mathbf{V}_{a}(\mathbf{x}) &=F_{V}^{a}(\{\mathbf{h}_{k}(\mathbf{x}) \oplus \mathbf{r}_{k}(\mathbf{x})\}_{k=1}^{m}) \\
\mathbf{a}(\mathbf{x}, \mathbf{d}) &=F^{a}\left(Att\left(\mathbf{Q}_{a}(\mathbf{x}), \mathbf{K}_{a}(\mathbf{x}), \mathbf{V}_{a}(\mathbf{x})\right)\right)
\end{aligned}
\end{equation}
where $F_{Q}^{a}$, $F_{K}^{a}$, $F_{V}^{a}$ denote the appearance layers producing the query, key, and value matrices $\mathbf{Q}_{a}(\mathbf{x})$, $\mathbf{K}_{a}(\mathbf{x})$, $\mathbf{V}_{a}(\mathbf{x})$, respectively. $F^{a}$ is the appearance feed-forward layer.

\subsection{Differential Rendering}
\label{Differential_Rendering}

After we get the geometric code and appearance code of any 3D point, we design a two-stage MLP network $F(\cdot)$ to build the neural feature field.
\begin{equation}
(\mathbf{f}, \sigma) = F(\mathbf{g}(\mathbf{x}), \mathbf{a}(\mathbf{x}, \mathbf{d}))
\end{equation}

Unlike classical NeRF\cite{NeRF}, which regresses color $\mathbf{c}$ and density $\sigma$, our decoder outputs the intermediate feature $\mathbf{f}$ and density $\sigma$. However, the original volume rendering process is memory-consuming, we use a combination of volume rendering and neural rendering to get the final image.

\begin{figure*}[ht]
    \centering
    \includegraphics[width=1.0\linewidth]{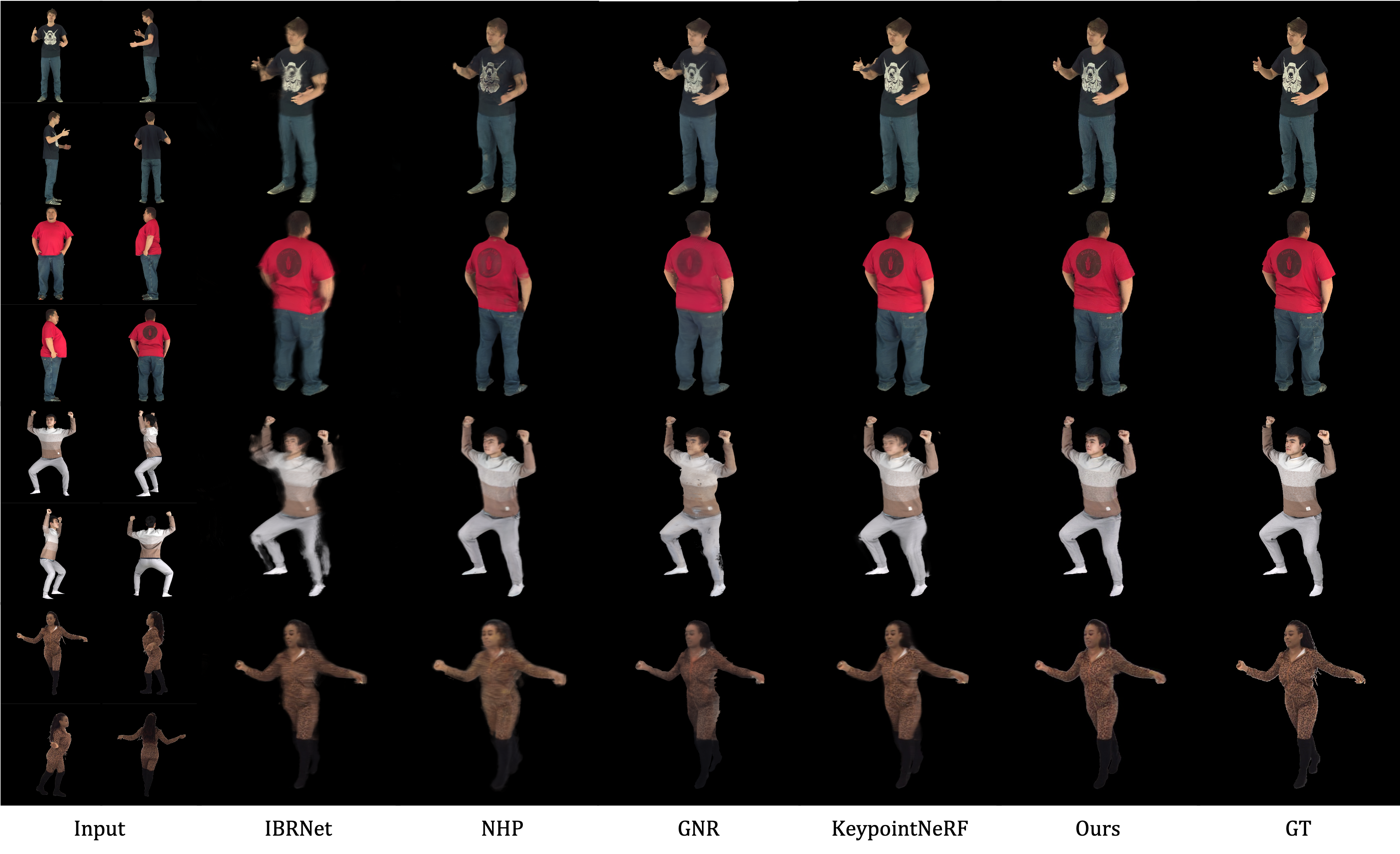}
    \vspace{-2.5em}
    \caption{{\bf Qualitative comparison with generalizable NeRFs}. We input $m = 4$ multi-view images of unseen identity, and our method produces a more photo-realistic novel view image compared to other state-of-the-art generalizable human NeRFs\cite{ibrnet, NHP, genebody, keypointNeRF}. The first two rows are from Multi-Garment\cite{Multi-Garment}, the third row from THuman2.0\cite{THuman2.0} and the last row from GeneBody\cite{genebody}.}
    \label{fig:results_on_GeNeRFs}
    \vspace{-0.5em}
\end{figure*}

\input{tables/generalizable_nerf}

{\bf 3D Volume Rendering}. We use the same volume rendering techniques as in NeRF~\cite{NeRF} to render the neural radiance field into a 2D image. Then the pixel colors are obtained by accumulating the colors and densities along the corresponding camera ray $\mathbf{\tau }$. In practice, the continuous integration is approximated by summation over sampled $N$ points $\left\{\mathbf{x}_{i}\right\}_{i=1}^{N}$ between the near plane and the far plane along the camera ray $\mathbf{\tau }$. 
\begin{equation}
\label{eq:volume rendering}
\begin{aligned}
\mathbf{\mathcal{F} }(\mathbf{\tau }) &= \sum_{i=1}^{N} \alpha_{i}\left(\mathbf{x}_{i}\right) \prod_{j<i}\left(1-\alpha_{j}\left(\mathbf{x}_{j}\right)\right) \mathbf{f}\left(\mathbf{x}_{i}\right) \\
\mathbf{\mathcal{M} }(\mathbf{\tau }) &= \sum_{i=1}^{N} \alpha_{i}\left(\mathbf{x}_{i}\right) \prod_{j<i}\left(1-\alpha_{j}\left(\mathbf{x}_{j}\right)\right) \\
\alpha_{i}(\mathbf{x}) &= 1-\exp \left(-\sigma(\mathbf{x}) \delta_{i}\right)  \\
\end{aligned}
\end{equation}
where $\delta_{i}=\left\|\mathbf{x}_{i+1}-\mathbf{x}_{i}\right\|_{2}$ is the distance between adjacent sampling points. $\alpha_{i}(\mathbf{x})$ is the alpha value for $\mathbf{x}$. The intermediate feature image $I_{\mathcal{F}}\in \mathbb{R}^{\frac{H}{2} \times \frac{W}{2} \times M_{\mathcal{F}}}$ and the silhouette image $I_{\mathcal{M}}\in \mathbb{R}^{\frac{H}{2} \times \frac{W}{2} \times 1}$ is obtained by \cref{eq:volume rendering}.

{\bf 2D Neural Rendering}.
\label{Neural_Rendering}
We utilize a 2D convolutional network $G_{\theta}$ to convert the intermediate feature image $I_{\mathcal{F}} \in \mathbb{R}^{\frac{H}{2} \times \frac{W}{2} \times M_{\mathcal{F}}}$ rendered by volume rendering into the final synthesized image $I_t \in \mathbb{R}^{H \times W \times 3}$.

\begin{equation}
    I_t \longleftarrow G_{\theta}\left( I_{\mathcal{F}} \right)
\end{equation}
where $\theta$ is the parameters of the 2D neural rendering network $G$, which means the rendering procedure is learnable.

\subsection{Loss Functions}
\label{Loss_Functions}

To stabilize the training procedure, we adopt the pixel-wise L2 loss widely used in \cite{PixelNeRF,NHP,genebody} to constrain  the rendered image $I_{t}$ and the alpha image $I_{\mathcal{M}}$.

\begin{equation}
\mathcal{L}=\lambda_{r}\left\|\tilde{I}_{t} - I_{t} \right\|_{2}^{2} + \lambda_{s}\left\|\tilde{I}_{\mathcal{M}} - I_{\mathcal{M}} \right\|_{2}^{2}
\end{equation}
where $\tilde{I}_{t}$, $\tilde{I}_{\mathcal{M}}$ are the ground-truth of the RGB image and silhouette image, respectively and $\lambda_{r}$, $\lambda_{s}$ are the weights. Beyond that, we also introduce the following loss functions to optimize the networks together,


{\bf Perceptual Loss}
\label{Perceptual_Loss}. We use a perceptual loss\cite{Perceptual_Loss} based on the VGG Network\cite{VGG}. It is more effective when the size of the images is closer to the network input, while it is memory intensive to render the whole image by volume rendering. To address these limitations, we adapt both neural rendering as well as partial gradient backpropagation.

\begin{equation}
    \mathcal{L}_{p}=\sum \frac{1}{N^{j}}\left|p^{j}\left(\tilde{I}_{t}\right)-p^{j}\left(I_{t}\right)\right|
\end{equation}
where $p^j$ is the activation function and $N^j$ is the number of elements of the $j$-th layer in the pretrained VGG network.




\begin{figure*}[ht]
    \centering
    \includegraphics[width=1.0\linewidth]{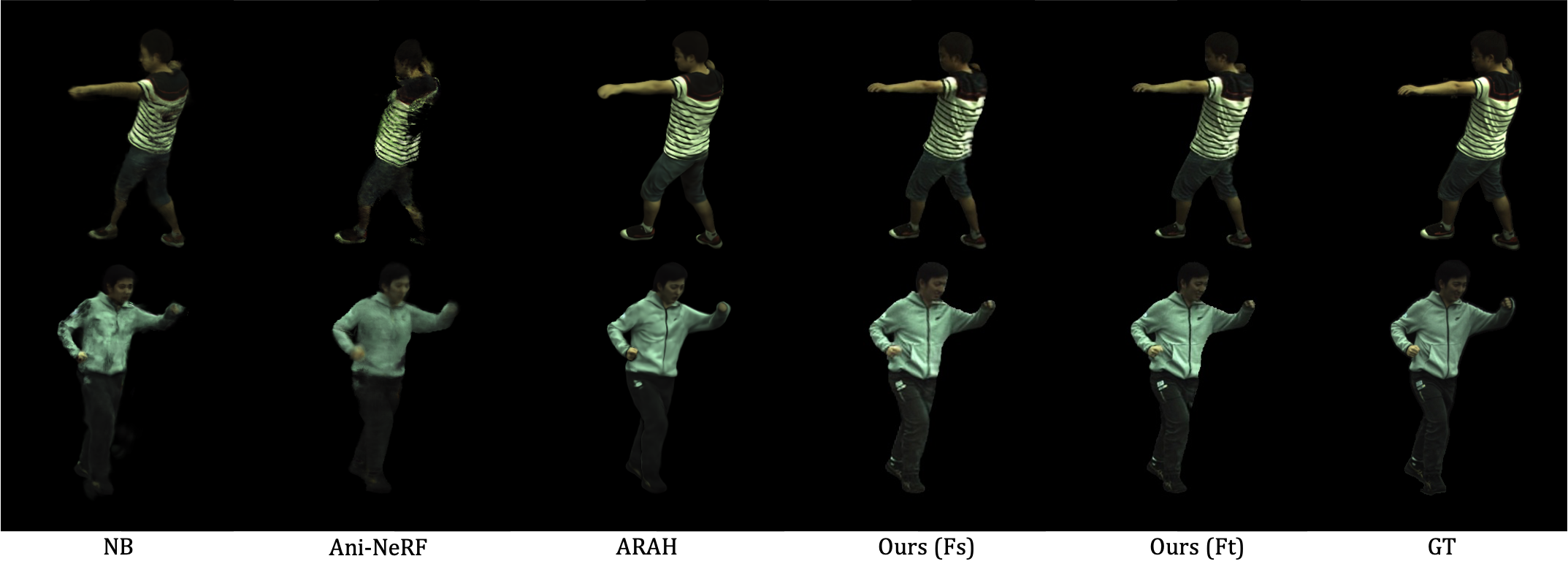}
    \vspace{-2.5em}
    \caption{{\bf Qualitative results of novel pose synthesis on ZJUMocap\cite{neuralbody} datasets}. Fs denotes training from scratch, Ft indicates fine-tuning the model after pretraining on Multi-Garment\cite{Multi-Garment} dataset.}
    \label{fig:novel_pose_on_zju}
    \vspace{-1.0em}
\end{figure*}

{\bf Normal Regularization}. Although NeRF\cite{NeRF} can produce realistic images, the geometric surfaces generated by Marching Cubes\cite{Marchingcubes} are extremely coarse and noisy. To alleviate it, we introduce normal regularization to constrain the normal among adjacent points.
\begin{equation}
\begin{aligned}
\mathcal{L}_{n}&=\sum_{\mathbf{x}_{s} \in \mathcal{S}}\left\|\mathbf{n}\left(\mathbf{x}_{s}\right)-\mathbf{n}\left(\mathbf{x}_{s}+\epsilon\right)\right\|_{2} \\
&\mathbf{n}\left(\mathbf{x}_{s}\right)=\frac{\nabla_{\mathbf{x}_{s} }\sigma \left(\mathbf{x}_{s}\right)}{\left\|\nabla_{\mathbf{x}_{s}}\sigma \left(\mathbf{x}_{s}\right)\right\|_{2}} 
\end{aligned}
\end{equation}
where $\mathcal{S}$ is the points set randomly sampled near the SMPL mesh surface. $\mathbf{n}\left(\mathbf{x}_{s}\right)$ is the normal of the sampled point $\mathbf{x}_{s}$ and $\epsilon$ is a gaussian random noise with a variance of $0.1$.

The final loss can be summarized as
\begin{equation}
\mathcal{L}_{full}= \mathcal{L} + \lambda_{p}\mathcal{L}_{p} + 
\lambda_{n}\mathcal{L}_{n}
\end{equation}
 where $\lambda_{p}$ and $\lambda_{n}$ are the weights of the perceptual loss and the normal regularization, respectively.

\subsection{Efficient Training}

\input{tables/persence_nerf}

\label{Efficient_Training}During training, we select $m$ multi-view images $\left\{I_{k}\right\}_{k=1}^{m}$ as inputs to build the generalizable model-based neural radiance fields and synthesize the target image $I_{t}$ with given camera pose. It is memory-consuming to synthesize the whole image at the same time by volume rendering, so we only generate an image patch of the resolution $H_{p}\times W_{p}$ sampled randomly from the whole target image, which means we only need to synthesize the half intermediate feature image of the resolution $\frac{H_{p}}{2} \times \frac{W_{p}}{2}$ by volume rendering. Meanwhile, since the perceptual loss requires a large enough image as input, we use partial gradient backpropagation introduced in CIPS-3D\cite{CIPS-3D} to further reduce the memory cost caused by volume rendering. Specifically, we randomly choose $n_{p}$ camera rays to participate in the gradient calculation, and the remaining rays $\frac{H_{p}}{2} \times \frac{W_{p}}{2} - n_{p}$ are not involved in gradient backpropagation.

%% file: tables/generalizable_nerf.tex
\begin{table*}[ht]
	\centering
	\scalebox{0.83}{
		\begin{tabular}{c|ccc|ccc|ccc|ccc}
			\toprule

			&\multicolumn{3}{c|}{Multi-Garment\cite{Multi-Garment}}
			&\multicolumn{3}{c|}{THuman2.0\cite{THuman2.0}}
                &\multicolumn{3}{c|}{ZJUMocap\cite{neuralbody}}
                &\multicolumn{3}{c}{GeneBody\cite{genebody}} 
			\\
			\midrule
			
			Model & PSNR$\uparrow$ & SSIM$\uparrow$ & LPIPS$\downarrow$   & PSNR$\uparrow$ & SSIM$\uparrow$ & LPIPS$\downarrow$ 
			& PSNR$\uparrow$ & SSIM$\uparrow$ & LPIPS$\downarrow$
                & PSNR$\uparrow$ & SSIM$\uparrow$ & LPIPS$\downarrow$\\
			\midrule
			
			IBRNet\cite{ibrnet} & 28.44 & 0.924 & 0.0917 & 25.66 & 0.916 & 0.1033 & 25.25 & 0.876 & 0.2323 & \bf{24.71} & 0.889 & 0.1364 \\
			
			NHP\cite{NHP} & 26.04 & 0.925 & 0.0701 & 26.99 & 0.935 & 0.0734 & 25.92 & 0.904 & 0.1623 & 22.75 & 0.872 & 0.1659\\
			
			GNR\cite{genebody} & 28.61 & 0.937 & 0.0511 & 25.82 & 0.929 & 0.0605 & 25.39 & 0.903 & 0.1306 & 22.21 & 0.887 & 0.1254\\
			
			KeypointNeRF\cite{keypointNeRF}  & 28.36 & 0.938 & 0.0471 & 25.93 & 0.929 & 0.0607 & 25.85 & 0.910 & 0.1092 & 24.34 & 0.902 & 0.1236\\
			
			Ours & \bf{30.18} & \bf{0.947} & \bf{0.0305} & \bf{28.88} & \bf{0.952} & \bf{0.0335} & \bf{26.74} & \bf{0.919} & \bf{0.0955} & 23.90 & \bf{0.906} & \bf{0.0865}\\
			
			\bottomrule
	\end{tabular}}
	\vspace{-0.5em}
	\caption{{\bf Quantitative comparisons with the generalizable NeRF methods}. We evaluate the novel view synthesis performance on the unseen identity of different datasets. Our method significantly outperforms the state-of-the-art methods.}
    \vspace{-1em}
	\label{tab:generalization}
\end{table*}

%% file: tables/persence_nerf.tex
\begin{table}[ht]
	\centering
	\scalebox{0.75}{
		\begin{tabular}{c|ccc|ccc}
			\toprule
			
			&\multicolumn{3}{c|}{Novel View Synthesis} 
			&\multicolumn{3}{c}{Novel Pose Synthesis}
			\\
			\midrule
			
			Model & PSNR$\uparrow$ & SSIM$\uparrow$ & LPIPS$\downarrow$   & PSNR$\uparrow$ & SSIM$\uparrow$ & LPIPS$\downarrow$ \\
			\midrule
			
			NB\cite{neuralbody} & 28.30 & 0.9462 &  0.0951 & 23.86 & 0.8971 & 0.1427 \\
			
			Ani-N\cite{animnerf_zju} & 26.19 & 0.9213 & 0.1399 & 23.38 & 0.8917 & 0.1594 \\
			A-NeRF\cite{a-nerf} & 27.43 & 0.9379 & 0.1019 & 22.40 & 0.8629 & 0.1991 \\
                ARAH\cite{arah} & \bf{28.51} & \bf{0.9483} & 0.0813 & 24.63 & 0.9112 & 0.1070 \\
                Ours (Fs) & 27.56 & 0.9314 & 0.0904 & 26.68 & 0.9241 & 0.0984 \\
			Ours (Ft) & 28.45 & 0.9419 & \bf{0.0733} & \bf{27.63} & \bf{0.9361} & \bf{0.0798} \\
			\bottomrule
	\end{tabular}}
	\vspace{-0.75em}
	\caption{{\bf Quantitative comparisons with case-specific optimization methods on ZJUMocap dataset.}
 }
    \vspace{-1.0em}
    \label{tab:novel_view_pose_on_zju}
\end{table}

%% file: sections/04_experiments.tex
\section{Experiments}
\subsection{Datasets}

We conduct experiments on two synthesized datasets Thuman2.0\cite{THuman2.0} and Multi-garment\cite{Multi-Garment} and real-world datasets Genebody\cite{genebody} and ZJUMocap\cite{neuralbody} for the generalizable scene task.
The Thuman2.0 dataset contains $525$ human scan meshes, of which we selected $400$ for training and the remaining $125$ for testing. For the Multi-garment dataset, we used $70$ meshes for training and $25$ meshes for evaluation. For each scanned mesh, we rendered it into $66$ multi-view images of resolution $1024\times 1024$. 
\begin{figure}[ht]
    \vspace{-0.5em}
    \centering
    \includegraphics[width=1.0\linewidth]{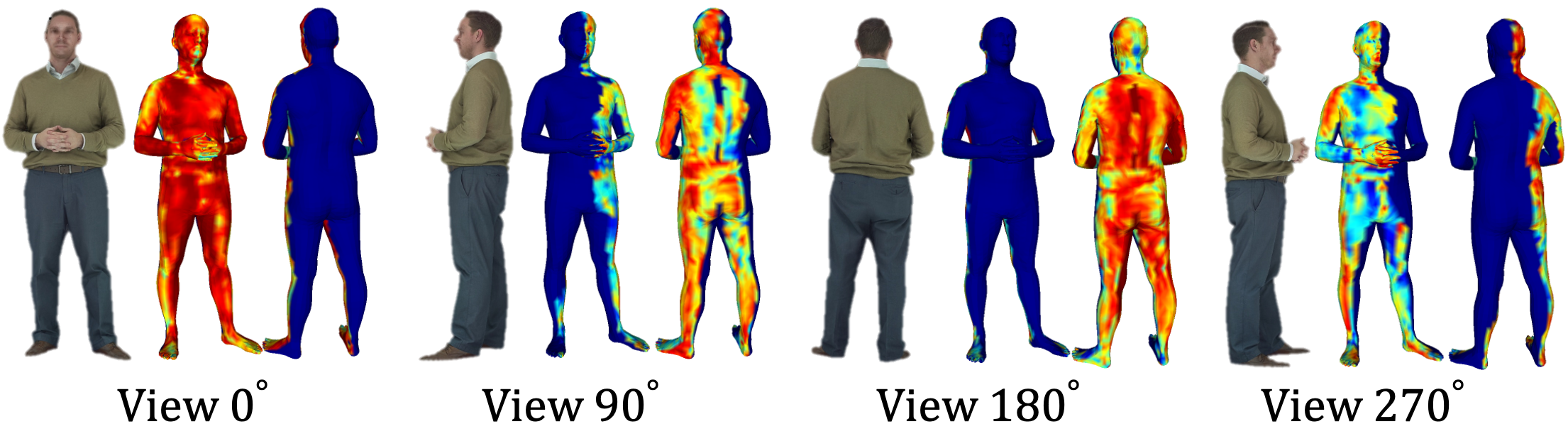}
    \vspace{-2.0em}
    \caption{
    {\bf The visualization of  visibility-based attention confidence}. We visualize the contribution of different input views to the SMPL vertices. (Red indicates high confidence, while blue represents low confidence.)
    }
    \vspace{-1.0em}
    \label{fig:att_vis}
\end{figure}
Specifically, we first place each scanned mesh into the center of a unit sphere at a distance of $5.4m$, with the camera orientation always pointing towards the center of the sphere. We move the camera around the sphere, sample a yaw angle from $0^{\circ}$ to $60^{\circ}$ with an interval of $30^{\circ}$, and sample a roll angle from
$0^{\circ}$ to $360^{\circ}$ with an interval of $30^{\circ}$. 
The Genebody consists of $50$ sequences at a $48$ synchronized cameras setting, each of which has $150$ frames.
Specifically, we choose $40$ sequences for training and another $10$ sequences for testing. For ZJUMocap, which captures $10$ dynamic human sequences with $21$ synchronized cameras, we use 7 sequences
for training and the rest 3 sequences for testing. 
To compare with the case-specific methods, we conduct experiments about novel view synthesis and novel pose synthesis on ZJUMocap. Following the evaluation protocols used in NB\cite{neuralbody}, we select $4$ fixed view videos for training.

\begin{figure*}[ht]
    \centering
    \includegraphics[width=1.0\linewidth]{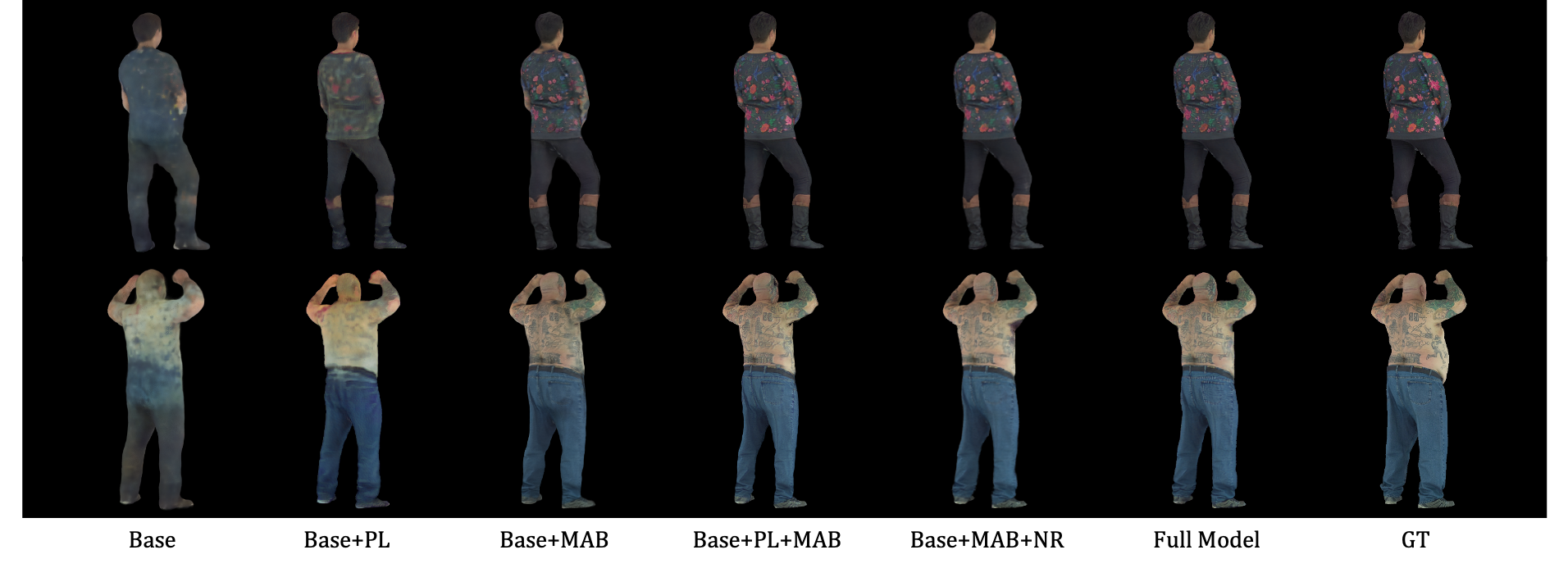}
    \vspace{-2.5em}
    \caption{{\bf Qualitative results of ablation studies on Multi-Garment dataset}.
    }
    \vspace{-1em}
    \label{fig:Alabtion_study}
\end{figure*}


\subsection{Evaluation Metrics}

We evaluate our method with state-of-the-art generalizable or per-scene optimized methods to verify the superiority of our performance. We formulate comparative experiments on both geometric reconstruction and novel view synthesis. 
\input{tables/ablation_study}
For quantitative comparison, we adopt peak signal-to-noise ratio (PSNR), structural similarity index (SSIM), and learned perceptual image patch similarity (LPIPS\cite{LPIPS}) to evaluate the similarity between the rendered image and the ground-truth. Meanwhile, we also adopt chamfer distance (Chamfer) and point-to-surface distance (P2S) for geometric quality evaluation.

\subsection{Implementation Details}

In our experiments, we choose $m = 4$ multi-view images $\left\{I_{k}\in \mathbb{R}^{512 \times 512 \times 3}\right\}_{k=1}^{m} $ as input to synthesize the target image $I_{t}\in \mathbb{R}^{512 \times 512 \times 3}$.
During training, the input multi-view images are selected randomly, while selected uniformly surrounding the person (\ie, the front, back, left, and right views) for evaluation.
The resolution of the patch image during training is $H_{p} = W_{p} = 224$. 
The SMPL parameters are obtained using EasyMocap\cite{neuralbody}. 
The size of our 3D feature volume $\mathcal{G}$ is $224^{3}$.
For partial gradient backpropagation, we randomly sample $n_{p} = 4,096$ camera rays from the target image patch to improve memory efficiency.
We then uniformly query $N = 64$ samples from our feature volume along the camera ray.
We train our network end-to-end by using the Adam\cite{adam} optimizer, and the base learning rate
is $5\times 10^{-4}$ which decays exponentially along with the optimization. We train $200,000$ iterations on two Nvidia RTX3090 GPUs with a batch size of $4$. The loss weights $\lambda _{r} = 1$, $\lambda _{s} = 0.1$, $\lambda _{p} = 0.01$, $\lambda _{n} = 0.01$.

\subsection{Evaluation.}

\textbf{Comparison with generalizable NeRFs}.
\begin{figure}[ht]
    \centering
    \includegraphics[width=1.0\linewidth]{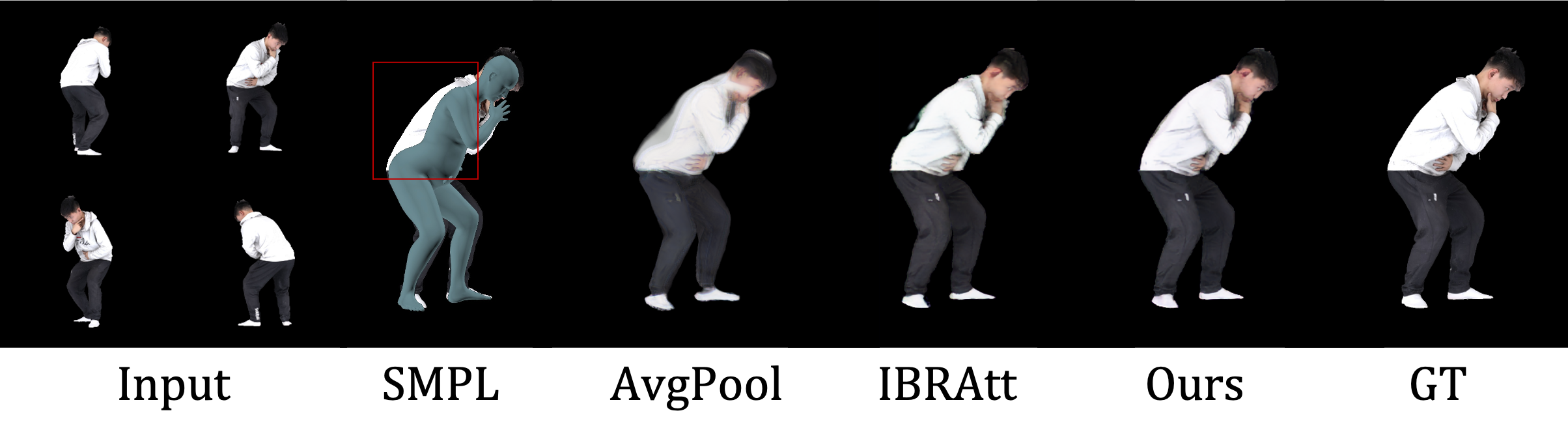}
    \vspace{-2.5em}
    \caption{{\bf Qualitative results of different multi-view fusion mechanisms}.}
    \label{fig:Ablation_on_attention}
    \vspace{-1.0em}
\end{figure}
\input{tables/ablation_on_attention.tex}
We compare our method with state-of-the-art generalizable methods IBRNet\cite{ibrnet}, NHP\cite{NHP}, GNR\cite{genebody} and KeypointNeRF\cite{keypointNeRF}.
We retrain all aforementioned networks with the official training protocols on GeneBody\cite{genebody}, Multi-Garment\cite{Multi-Garment}, and THuman2.0\cite{THuman2.0} datasets. Specially, we also use $m=3$ views as input on ZJUMocap\cite{neuralbody} dataset following the evaluation protocol used in KeypointNeRF. The result can be seen in \cref{tab:generalization} and \cref{fig:results_on_GeNeRFs}, which shows our method generalizes to unseen identities well and outperforms the methods compared. IBRNet, which learns a general view interpolation function to synthesize the novel view from a sparse set of nearby views, is able to render high-fidelity images for views close to the input views while having very poor generalization for views far from the input views. Our method has better generalization of novel view synthesis and generates higher quality geometry due to the use of the geometry prior SMPL. KeypointNeRF utilizes sparse 3D keypoints
\begin{figure}[ht]
    \centering
    \includegraphics[width=1.0\linewidth]{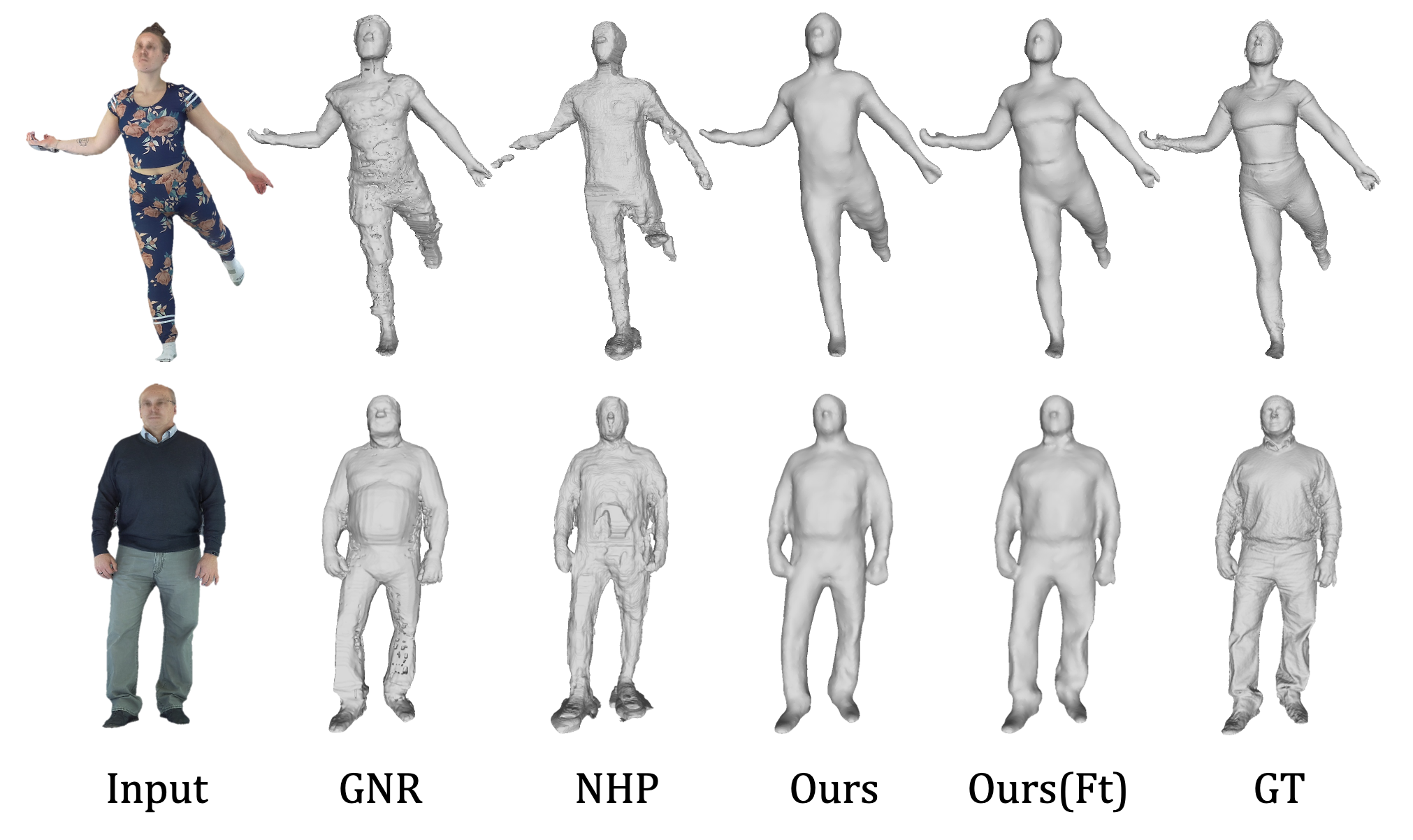}
    \vspace{-2.5em}
    \caption{{\bf Visualization results of 3D geometry reconstruction compared with different methods}.}
    \label{fig:geometry_results}
    \vspace{-0.5em}
\end{figure}
\input{tables/geometry}
as pose priors and has weak expressiveness for unseen poses when the pose diversity in the training set is insufficient. In our experiment, we choose the 3D joints of SMPL as the input of KeypointNeRF.
Compared to NHP and GNR, although we both employ SMPL as the geometry prior and suffer from inaccurate SMPL estimation, our method can alleviate the ambiguity of misalignment between geometry and pixel-aligned appearance. Meanwhile, benefiting from perceptual loss, our generated images have more photo-realistic 
 local details.
 For 3d reconstruction, the mesh surface extracted by Marching Cubes is smoother and more precise due to normal regularization compared with others as shown in \cref{fig:geometry_results} and \cref{tab:geometry}.

\textbf{Comparison with case-specific Methods}. We also compare with per-scene optimization methods
NB\cite{neuralbody}, Ani-NeRF\cite{animnerf_zju}, A-NeRF\cite{a-nerf}, ARAH\cite{arah}.
NB optimizes a set of structured latent codes associated with SMPL vertices, which are diffused into the observation space by using SparseConvNet. Since the 3D convolution in SparseConvNet is not rotation-invariant, NB has poor generalization on out-of-distribution poses. Ani-NeRF learns a backward LBS network to warp the observation space into the canonical space, which is not sufficient to model non-rigid deformations in complex poses. A-NeRF uses skeleton-relative embedding to model pose dependence deformation, which requires seeing the subjects from all views in varying poses. ARAH uses iterative root-finding for simultaneous ray-surface intersection search and correspondence search, which generalizes well to unseen poses.
As shown in Tab~\ref{tab:novel_view_pose_on_zju}, the performance of novel view synthesis is comparable with these methods, and it is reasonable since our network has more parameters($13.6$M) and struggles with overfitting when the training data is so limited without any pretraining. After pretraining on the Multi-Garment and finetuning $5,000$ steps on ZJUMocap, our results achieve a noticeable improvement.
Anyway, our method has superior generalization on novel pose synthesis, which is a more challenging task. Our results are more photorealistic and preserve more details like wrinkles and patterns as shown in Fig~\ref{fig:novel_pose_on_zju}, which benefit from the sparse multi-view input. 
 



\subsection{Ablation studies}

The baseline ({\bf Base}) is an extended version of NB
to express arbitrary identities as our baseline. Specifically, our structured latent codes are obtained by fusing multi-view input, rather than optimizing from scratch for a specific identity.
Beyond that, we introduce multi-view appearance blending ({\bf MAB}), perception loss ({\bf PL}), and neural rendering ({\bf NR}). The experimental results prove the effectiveness of each component as shown in \cref{tab:albation_study} and \cref{fig:Alabtion_study}. In addition, we explore the effects of different multi-view fusion mechanisms, and the experiments prove that our proposed visibility-based attention and geometry-guided attention are more effective than AvgPool\cite{PIFu, PixelNeRF} and IBRAtt\cite{ibrnet}.

%% file: tables/ablation_study.tex
\begin{table}[ht]
	\centering
	\scalebox{0.85}{
		\begin{tabular}{c|ccc}
			\toprule
			Model & PSNR$\uparrow$ & SSIM$\uparrow$ & LPIPS$\downarrow$  \\
			\midrule
			
			Base & 25.22 & 0.895 & 0.1048   \\
			
			Base+MAB & 27.08  & 0.915 & 0.0611 \\
			
			Base+PL & 27.75 & 0.913 & 0.0673 \\
			
			Base+MAB+PL & 28.72 & 0.929 & 0.0423 \\

            Base+MAB+NR & 30.03 & 0.940 & 0.0562 \\
   
			Full Model & \bf{30.18} & \bf{0.947} & \bf{0.0305}  \\
			\bottomrule
	\end{tabular}}
	\vspace{-0.5em}
	\caption{{\bf Quantitative results 
 of ablation studies on the Multi-garment}. Impact of the different components in our method.}
    \vspace{-0.75em}
	\label{tab:albation_study}
\end{table}

%% file: tables/ablation_on_attention.tex
\begin{table}[ht]
	\centering
	\scalebox{0.85}{
		\begin{tabular}{c|ccc}
			\toprule
			Model & PSNR$\uparrow$ & SSIM$\uparrow$ & LPIPS$\downarrow$  \\
			\midrule
			
			Ours with AvgPool & 27.81 & 0.9317 & 0.05059   \\
			
			Ours with IBRAtt & 28.50  & 0.9345 & 0.03413 \\
			
			Ours & \bf{28.88} & \bf{0.9518} & \bf{0.03349} \\
			\bottomrule
	\end{tabular}}
	\vspace{-0.5em}
	\caption{{\bf Quantitative results of different multi-view fusion mechanisms on the THuman2.0 dataset}. AvgPool is used in PIFu\cite{PIFu} and PixelNeRF\cite{PixelNeRF}. IBRAtt is proposed by IBRNet.}
	\label{tab:albation_on_attention}
    \vspace{-1.0em}
\end{table}

%% file: tables/geometry.tex
\begin{table}[ht]
	\centering
	\scalebox{0.9}{
		\begin{tabular}{c|cc|cc}
			\toprule
			
			&\multicolumn{2}{c|}{Multi-Garment\cite{Multi-Garment}} 
			&\multicolumn{2}{c}{THuman2.0\cite{THuman2.0}}
			\\
			\midrule
			
			Model & Chamfer$\downarrow$ & P2S$\downarrow$ & Chamfer$\downarrow$   & P2S$\downarrow$ \\
			\midrule
			
			GNR\cite{genebody} & 1.3570 & 1.8981 & 1.7899 & 2.5932  \\
			
			NHP\cite{NHP} & 1.4646 & 2.2438 & 1.6027 & 2.3921 \\
			
			Ours & 0.7175 & 0.6919 & 0.7444 & 0.6600  \\
			
			Ours(Ft) & \bf{0.3721} & \bf{0.3676} & \bf{0.5172} & \bf{0.4506}  \\
			\bottomrule
	\end{tabular}}
    \vspace{-0.5em}
	\caption{{\bf Quantitative comparisons of 3D geometry reconstruction}. Our method consistently outperforms other methods, capturing more local details after fine-tuning.}
    \vspace{-1.0em}
	\label{tab:geometry}
\end{table}

%% file: sections/05_limitations.tex
\section{Limitations}

There are some limitations of our method that need to be improved: i) Due to the minimal-clothed topology of SMPL, our model struggles to express extremely loose clothes and accessories. ii) When the testing pose is out-of-distribution, our method may produce some artifacts in results since 3D convolution in SparseConvNet\cite{spconv} is not rotation-invariant. iii) As the target view moves further away from the input views, artifacts tend to emerge in the unobserved areas.

%% file: sections/06_conclusion.tex
\section{Conclusion}


In this paper, we propose an effective framework to build generalizable model-based neural radiance fields (GM-NeRF) from sparse calibrated multi-view images of arbitrary performers. 
To improve generalization on novel poses and identities, we introduce SMPL as the structured geometric body embedding. However, inaccurate estimations of SMPL have a negative impact on the reconstruction results. 
To address this, we propose a novel geometry-guided multi-view attention mechanism that can effectively alleviate the misalignment between SMPL geometric prior feature and pixel-aligned feature.
Meanwhile, we propose strategies such as neural rendering and partial gradient backpropagation to efficiently train our network using perceptual loss. Extensive experiments show that our method outperforms concurrent works.